\documentclass{article}
\usepackage{spconf,amsmath,graphicx,hyperref}

\usepackage{booktabs}
\usepackage{multirow}
\usepackage{adjustbox}
\usepackage{amssymb}

\title{JAVIS: Joint Audio-Visual Interaction and Streaming \\for Real-Time Conversational Agents}
\name{Junseok Ahn$^1$, Jaehyun Park$^1$, Jeongsoo Choi$^1$, Joon Son Chung$^1$, Ji-Hoon Kim$^2$\thanks{Project page: \url{https://mm.kaist.ac.kr/projects/JAVIS}}}
\address{$^1$Korea Advanced Institute of Science and Technology, South Korea \\
$^2$Chung-Ang University, South Korea}
\begin{document}
\ninept
\maketitle
\begin{abstract}
Human communication is inherently multimodal, conveying intent not only through linguistic and vocal cues but also through visual signals.
However, existing conversational systems have mainly focused on text or speech, leaving out the visual feedback that is essential for natural interaction.
In this paper, we propose \textbf{JAVIS}, an end-to-end streaming framework for real-time audio-visual conversational agents.
Our key idea is to let the voice and the face speak from a single representation.
On top of a shared LLM, a Joint Refined Head produces a unified audio-visual representation, which is fed into dedicated streaming branches for speech and face generation.
This end-to-end design avoids intermediate bottlenecks, enabling real-time generation while keeping speech and face tightly synchronized.
In addition, we present \textbf{JAVIS-Syn}, a 334-hour synthetic corpus designed to address the lack of expressive audio-visual dialogue data, covering emotionally rich multi-turn conversations and assistant-style exchanges.
Experiments show that JAVIS achieves approximately 200$\times$ lower initial latency than a strong baseline, while outperforming baselines in lip synchronization and facial expression appropriateness.
\end{abstract}

\section{Introduction}
\label{sec:intro}

In our daily lives, we see the face of the person we are talking to and hear their voice.
As such, human communication is inherently multimodal~\cite{mcgurk1976hearing,kendon1967some,ahn2024slowfast,jang2026deep}.
Speakers convey intent not only through words and speech but also through facial expressions, head movements, and gaze.
Conversational AI systems, however, have evolved largely along the linguistic and auditory axes.
From text-based chatbots built on large language models (LLMs) to recent spoken dialogue models that directly understand and generate speech~\cite{team2025fun}, the medium of interaction has remained confined to text or speech.
The absence of a visual channel limits the expressiveness and naturalness of these systems, calling for audio-visual conversational agents that can respond with both voice and face in real time.

A common workaround is to cascade a conversational model with a separate generator, either a text LLM followed by text-to-audio-visual generation~\cite{ye2026talker, jang2024faces}, or a spoken dialogue system followed by speech-driven talking face generation~\cite{yang2025infinitetalk,ki2025float}.
However, such pipelines are inherently limited by their sequential structure, in which the intermediate modality acts as an information bottleneck.
Text discards prosodic nuance, and synthesized speech conveys only acoustic cues, preventing the downstream face generator from accessing the dialogue context encoded in the upstream model.
Furthermore, the sequential dependency between modules causes latency to accumulate, and errors propagate across stages and degrade audio-visual synchronization~\cite{kim2025tavid}.

\begin{figure}[t]
    \centering
    \includegraphics[width=.99\linewidth]{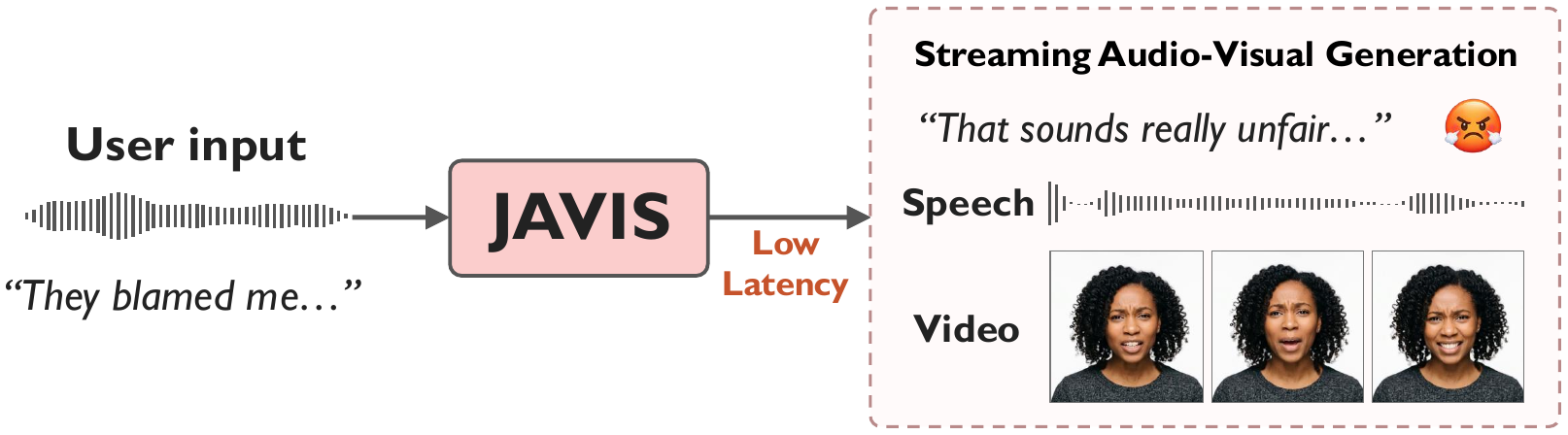}
    \caption{Overview of JAVIS. Given user speech, JAVIS jointly generates an audio-visual response in real time.}
    \label{fig:teaser}
\end{figure}

To date, only a few recent works~\cite{xie2025x,zhang2026ex} have explored end-to-end audio-visual dialogue generation.
X-Streamer~\cite{xie2025x} adopts a Thinker--Actor architecture, where a frozen speech LLM guides an Actor that generates interleaved text and speech tokens and synthesizes video latents via chunk-wise autoregressive diffusion.
Ex-Omni-2D~\cite{zhang2026ex} instead conditions a diffusion-based video generator on an LLM-generated visual plan and speech units produced by a separate speech generator.
However, both approaches rely on iterative generation of spatiotemporal video latents, which imposes a heavy computational burden on visual synthesis.
Real-time streaming generation of realistic visual responses thus remains an open problem, made even more challenging by the scarcity of expressive audio-visual dialogue data.

In this paper, we propose JAVIS, an end-to-end streaming framework for real-time audio-visual conversational agents.
Given a spoken query, JAVIS responds with synchronized speech and facial video in a streaming manner, as illustrated in Fig.~\ref{fig:teaser}.
Rather than animating the face after speech is synthesized, our framework jointly generates both modalities from a single representation.
Specifically, a Joint Refined Head produces a unified audio-visual representation, which is fed into two streaming branches for speech decoding and facial motion generation.
By bypassing intermediate text or speech, this design preserves the conversational intent in both modalities and maintains tight audio-visual synchronization.
Furthermore, to address the scarcity of audio-visual dialogue data, we construct \textbf{JAVIS-Syn}, a 334-hour synthetic corpus comprising emotionally rich multi-turn conversations and assistant-style exchanges.
Experiments demonstrate that JAVIS generates high-quality audio-visual responses in real time,
outperforming existing works in audio-visual synchronization.
Notably, our model reduces initial latency by approximately 200$\times$ compared with a strong baseline while maintaining comparable visual quality.
Human evaluation further confirms that JAVIS produces natural audio-visual responses with accurate lip synchronization and situation-appropriate facial expressions.

\section{Method}

\begin{figure}[t]
    \centering
    \includegraphics[width=.99\linewidth]{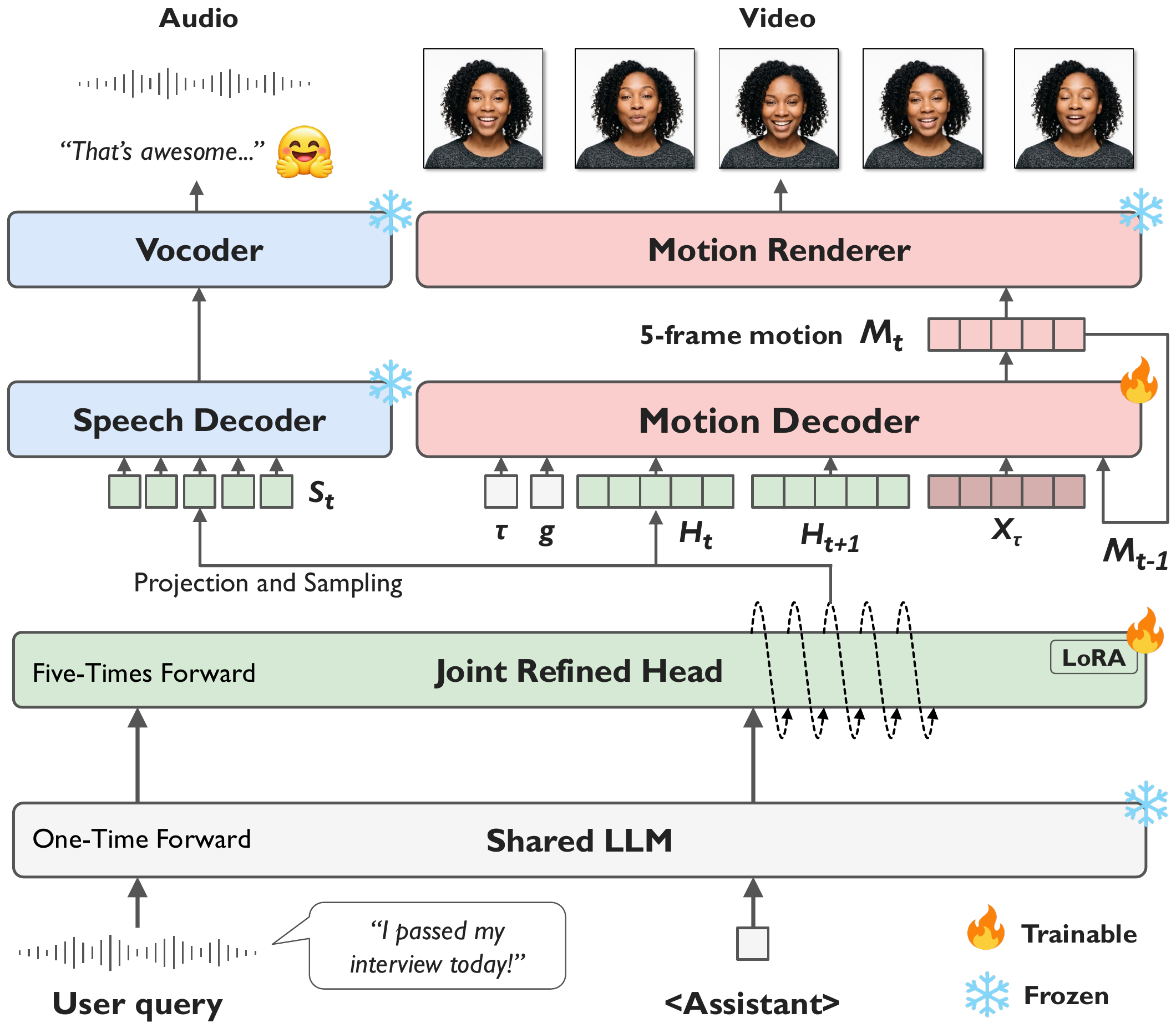}
    \caption{
    Network architecture.
    The Joint Refined Head produces a unified audio-visual representation $H_t$, which is fed into parallel speech and motion branches for streaming generation.
    The speech branch samples five speech tokens $S_t$ from $H_t$, while the Motion Decoder denoises $X_\tau$ at timestep $\tau$ to generate the five-frame motion chunk $M_t$, conditioned on $H_t$, $H_{t+1}$, the previous motion chunk $M_{t-1}$, and the reference facial state $g$.
    }
    \label{fig:overview}
\end{figure}

Given a spoken query, the dialogue history, and a reference portrait, JAVIS generates a spoken response along with a synchronized facial video of the reference identity, as depicted in Fig.~\ref{fig:overview}.
JAVIS is built upon the pre-trained Fun-Audio-Chat~\cite{team2025fun}, which pairs a 5~Hz shared LLM with a 25~Hz speech refined head for efficient streaming generation.
Rather than conditioning facial motion on the generated speech waveform, we repurpose the speech refined head into the Joint Refined Head (JRH) via LoRA~\cite{hu2022lora}, so that the JRH hidden states serve as a unified audio-visual representation while preserving the pre-trained speech generation capability.
The unified representation is fed into two parallel branches: a speech branch that synthesizes the spoken response, and a motion branch where a Motion Decoder autoregressively generates chunks of continuous facial motion, which are subsequently rendered by the Motion Renderer~\cite{ki2025float} with the reference portrait.
As verified in Sec.~\ref{sec:exprst}, this end-to-end architecture enables real-time streaming generation of situation-appropriate responses with tight audio-visual synchronization.

\subsection{Joint Refined Head}
The JRH repurposes the speech refined head of Fun-Audio-Chat~\cite{team2025fun} to produce a unified audio-visual representation for both speech token and facial motion generation.
Since the shared LLM and the JRH operate at 5~Hz and 25~Hz, respectively, the $t$-th shared LLM output is ungrouped into five conditioning vectors.
The JRH then processes the five conditioning vectors autoregressively, each combined with the embedding of the preceding speech token.
At each autoregressive step, the next speech token is sampled from a linear projection of the final-layer hidden state.
The resulting five hidden states are stacked into $H_t\in\mathbb{R}^{5\times d}$, where $d$ denotes the JRH hidden dimension, and serve as the unified representation for the $t$-th chunk.
The Motion Decoder takes
$H_t$ directly as its condition, bypassing waveform decoding and re-encoding for motion generation.

Throughout response generation, the JRH attends to the shared LLM outputs and dialogue history, allowing $H_t$ to encode the dialogue context beyond mere speech content.
To exploit this context while preserving the pre-trained speech generation capability, we freeze the shared LLM and jointly train the LoRA-adapted JRH and the Motion Decoder.
The JRH is thus supervised by both the cross-entropy loss $\mathcal{L}_{\mathrm{speech}}$ for speech token prediction and the motion generation loss $\mathcal{L}_{\mathrm{motion}}$ (Sec.~\ref{subsec:branches}), encouraging $H_t$ to capture facial dynamics that are synchronized with speech.

\subsection{Streaming Audio-Visual Generation}
\label{subsec:branches}
\noindent{\bf Speech branch.}
The speech branch adopts the pre-trained speech decoder and vocoder of CosyVoice~3~\cite{du2025cosyvoice}, which are kept frozen throughout training.
The flow-matching speech decoder converts the sampled speech tokens into mel-spectrograms, which the vocoder then synthesizes into waveforms.
The decoder's chunk-aware causal design enables incremental decoding, so the spoken response can be streamed while subsequent speech tokens are still being generated.

\noindent\textbf{Motion branch.}
The motion branch consists of the Motion Decoder, which autoregressively generates facial motion chunk by chunk from the unified representation, and the Motion Renderer~\cite{ki2025float}, which renders the generated motion with the reference portrait into video frames.
We represent the facial motion of each frame as 20-dimensional coefficients over the orthonormal motion basis of FLOAT~\cite{ki2025float}.
We group the coefficients of five consecutive frames into a single motion chunk $M_t\in\mathbb{R}^{5\times 20}$, which spans 0.2 seconds at 25 FPS.
This grouping makes each motion chunk correspond to $H_t$, so that each motion frame is naturally aligned with a speech token, keeping the motion branch synchronized with the speech branch at 25~Hz.
Since lip movements are often affected by upcoming phonetic context~\cite{chandrasekaran2009natural}, the Motion Decoder also takes $H_{t+1}$ as a one-chunk lookahead.
It further receives the previous motion chunk $M_{t-1}$ for temporal continuity and a reference facial state $g$ extracted from the reference portrait.
The resulting condition is $c_t=(H_t, H_{t+1}, M_{t-1}, g)$.

Conditioned on $c_t$, the Motion Decoder generates $M_t$ with a bidirectional Transformer.
It takes the concatenated sequence of $c_t$, the noisy motion chunk $X_\tau$, and the timestep embedding $\tau$, and is trained with the conditional flow matching objective~\cite{lipman2023flow}:
\begin{equation}
\mathcal{L}_{\mathrm{motion}}
=\mathbb{E}\!\left[\left\|v_\theta(X_\tau,\tau;c_t)-v\right\|^2\right],
\end{equation}
where $v_\theta$ denotes the Motion Decoder and $v$ is the target \mbox{velocity}.
The overall objective is $\mathcal{L}=\mathcal{L}_{\mathrm{speech}}+\mathcal{L}_{\mathrm{motion}}$, under which the LoRA-adapted JRH and the Motion Decoder are optimized jointly.
At inference, once $H_{t+1}$ is available, $\hat{M}_t$ is sampled from Gaussian noise with an Euler ODE solver and used as the previous motion chunk for the next chunk, with the first chunk conditioned on zero motion.
The generated motion is then rendered by the Motion Renderer with the reference portrait and streamed together with the corresponding speech segments.
By operating directly on $H_t$ with low-dimensional facial motion, the Motion Decoder achieves low-latency generation, while its fixed input window keeps the per-chunk cost independent of response length.

\begin{table*}[!t]
\centering
\caption{
Objective quality of audio-visual responses,
evaluated in terms of audio-visual synchronization, video quality,
and speech quality.
Arrows indicate the preferred direction.
Best and second-best results are shown in
\textbf{bold} and \underline{underlined}, respectively.
}
\label{tab:objective_quality}
\vspace{2mm}

\small
\setlength{\tabcolsep}{4pt}
\renewcommand{\arraystretch}{1.12}

\begin{tabular*}{\textwidth}{
    @{\extracolsep{\fill}}lcccccccc@{}
}
    \toprule
    \multirow{2}{*}{Method}
    & \multicolumn{2}{c}{AV sync}
    & \multicolumn{3}{c}{Video quality}
    & \multicolumn{3}{c}{Speech quality} \\
    \cmidrule(lr){2-3}
    \cmidrule(lr){4-6}
    \cmidrule(lr){7-9}
    & LSE-C $\uparrow$
    & LSE-D $\downarrow$
    & IQ $\uparrow$
    & SC $\uparrow$
    & MS $\uparrow$
    & WER (\%) $\downarrow$
    & UTMOS $\uparrow$
    & SpkSIM $\uparrow$ \\
    \midrule

    FAC (s2s) + FLOAT
    & 7.94
    & 8.26
    & 0.743
    & 0.989
    & 0.9928
    & \underline{0.41}
    & 4.20
    & \underline{0.689} \\

    FAC (s2s) + InfiniteTalk
    & \textbf{9.59}
    & \underline{6.86}
    & \textbf{0.754}
    & 0.991
    & 0.9936
    & \underline{0.41}
    & 4.20
    & \underline{0.689} \\

    FAC (s2t) + Talker-T2AV
    & \underline{8.72}
    & 7.04
    & 0.738
    & \textbf{0.994}
    & \textbf{0.9954}
    & 0.54
    & 3.79
    & 0.575 \\

    Ex-Omni-2D
    & 5.24
    & 11.06
    & 0.734
    & 0.991
    & \underline{0.9947}
    & \textbf{0.31}
    & \textbf{4.32}
    & 0.525 \\

    \midrule
    \textbf{JAVIS}
    & \textbf{9.59}
    & \textbf{6.74}
    & \underline{0.744}
    & \underline{0.993}
    & 0.9935
    & 0.44
    & \underline{4.29}
    & \textbf{0.701} \\
    \bottomrule
\end{tabular*}

\end{table*}

\begin{table}[!t]
\centering
\caption{
Statistics of JAVIS-Syn after lip-sync filtering.
$^\dagger$~For VoiceAssistant-400K, only assistant responses are rendered as videos;
the duration excludes the corresponding user-query audio.
}
\label{tab:dataset_statistics}
\vspace{2mm}

\small
\setlength{\tabcolsep}{5pt}
\renewcommand{\arraystretch}{1.12}

\begin{adjustbox}{max width=\columnwidth}
\begin{tabular}{@{}lrrr@{}}
    \toprule
    Dialogue Source & \# Dialogues & \# Video Utterances & Duration (h) \\
    \midrule
    DailyTalk & 2,006 & 18,400 & 20.10 \\
    ShareChatX-Emotion & 26,169 & 195,908 & 263.89 \\
    VoiceAssistant-400K$^\dagger$ & 11,472 & 11,472 & 50.13 \\
    \midrule
    \textbf{Total} & \textbf{39,647} & \textbf{225,780} & \textbf{334.13} \\
    \bottomrule
\end{tabular}
\end{adjustbox}

\end{table}

\subsection{JAVIS-Syn}
Training audio-visual conversational systems requires emotionally rich, multi-turn dialogues in which each spoken response is paired with an expressive talking face video.
However, existing audio-visual dialogue corpora~\cite{park2024let, agrawal2025seamless} either exhibit a limited range of facial expressions or focus only on human--human conversations without covering the user--assistant interaction format targeted by conversational agents.
To address this, we construct JAVIS-Syn, a 334-hour synthetic audio-visual dialogue dataset that provides expressive facial motion and assistant-style interactions.

We utilize three dialogue sources to construct JAVIS-Syn.
DailyTalk~\cite{lee2023dailytalk} and ShareChatX-\allowbreak Emotion~\cite{cheng2025omnichat} provide multi-turn conversations with utterance-level emotion labels, while VoiceAssistant-\allowbreak 400K~\cite{chen2025slam}\footnote{Since the released answer speech is synthesized in a single voice, we resynthesize it with CosyVoice~3~\cite{du2025cosyvoice} using reference speech from EARS~\cite{richter2024ears}, and assign a neutral emotion label to all answers.} provides question--answer pairs reflecting how an AI assistant responds to user requests.
For DailyTalk and ShareChatX-Emotion, we render every utterance in each dialogue using its emotion label, whereas for VoiceAssistant-400K, we retain the user query as audio and render only the assistant response with a neutral emotion prompt.
The corresponding videos are generated by InfiniteTalk~\cite{yang2025infinitetalk}, an audio-driven talking face model whose facial expressions can be controlled through an emotion prompt.
Reference portraits are drawn from a pool of 1,220 frontal, neutral-expression portraits generated with Gemini 3.1 Flash Image~\cite{googledeepmind2026gemini31flashimage}, covering diverse genders and ages.
Each speaker is assigned a portrait matching their gender and age group, obtained from the speaker metadata, and the same portrait is used for all utterances of that speaker within a dialogue.
The generated videos are filtered by lip-sync quality using SyncNet~\cite{chung2016out}, yielding 225,780 utterance videos totaling 334.13 hours, with the resulting dataset composition summarized in Table~\ref{tab:dataset_statistics}.
We will publicly release the dataset upon acceptance.

\section{Experiments}
\label{sec:exprst}
\subsection{Experimental Setup}

\noindent\textbf{Training.}
JAVIS is trained in two stages.
In the first stage, we train the model in a monologue setting using 940,497 single-utterance clips from the Talker-T2AV~\cite{ye2026talker} dataset.
The model is prompted to repeat the input speech, and the same speech together with its facial motion serves as the training target.
In the second stage, we adapt the model to dialogue response generation using 716,565 samples from MultiDialog~\cite{park2024let}, JAVIS-Syn, and a subset of Seamless Interaction~\cite{agrawal2025seamless}.
Each sample contains up to three preceding turns as context, and the loss is applied only to the final assistant response.

\noindent\textbf{Implementation details.}
In both stages, only the LoRA adapter of the JRH and the Motion Decoder are trained, while the remaining modules are frozen.
The Motion Decoder is an eight-layer Transformer with a hidden dimension of 1,024, and LoRA is applied with rank 16 and $\alpha=32$.
Both stages use AdamW with a learning rate of $10^{-4}$, a cosine schedule, and a batch size of 8.
At inference, motion is sampled with 10 Euler steps and a noise temperature of 0.7.

\noindent\textbf{Evaluation.}
JAVIS is compared with an end-to-end system, Ex-Omni-2D\footnote{We use the publicly released Teacher checkpoint.}, and three cascaded systems that pair Fun-Audio-Chat (FAC) with FLOAT~\cite{ki2025float}, InfiniteTalk~\cite{yang2025infinitetalk}\footnote{We use the four-step distilled configuration.}, or Talker-T2AV~\cite{ye2026talker}.
Since FAC produces both a text and a speech response, FLOAT and InfiniteTalk animate its speech output (s2s), whereas Talker-T2AV generates both speech and video from its text output (s2t).
All models are evaluated on 140 samples from VStyle~\cite{zhan2026vstyle}, which is unseen during training.

\noindent\textbf{Metrics.}
For video, we measure image quality (IQ), subject consistency (SC), and motion smoothness (MS) using VBench~\cite{huang2024vbench}.
Lip synchronization is measured with LSE-C and LSE-D~\cite{chung2016out}.
For speech, we measure speech naturalness (UTMOS~\cite{saeki2022utmos}), speaker similarity (SpkSIM~\cite{chen2022wavlm}), and intelligibility (WER~\cite{radford2023robust}) of the generated speech.
Generation efficiency is evaluated by the initial latency, defined as the delay until the first audio and video outputs are both available, and the real-time factor (RTF), computed as the ratio of total generation time to the duration of the generated response.

\begin{table}[t]
    \centering
    \caption{Streaming latency on two RTX 4090 GPUs, averaged over five queries.
        Latency: time until both the first audio and video packets are available.
        RTF: wall-clock time divided by the length of the generated response.
        Len.: the length of the generated response.}
    \label{tab:streaming}
    \vspace{2mm}
    \setlength{\tabcolsep}{4pt}
    \begin{tabular}{@{}lccc@{}}
        \toprule
        System & Latency (s) $\downarrow$ & RTF $\downarrow$ & Len. (s) \\
        \midrule
        FAC + FLOAT                & 6.73    & 0.569 & 11.8 \\
        FAC + FLOAT (2\,s chunks)  & 2.09    & 0.560 & 11.8 \\
        FAC + InfiniteTalk         & 252.0   & 20.9  & 11.8 \\
        FAC + Talker-T2AV          & 113.3   & 10.8  & 10.5 \\
        Ex-Omni-2D                 & 1210.7  & 116.6 & 10.5 \\
        \midrule
        JAVIS                      & \textbf{1.27} & \textbf{0.541} & 12.1 \\
        \bottomrule
    \end{tabular}
\end{table}

\subsection{Objective Evaluation}
\label{sec:objective}

\noindent\textbf{Audio-visual quality.}
We compare JAVIS against cascaded and end-to-end baselines, and the results are shown in Table~\ref{tab:objective_quality}.
JAVIS achieves the most accurate audio-visual synchronization, with the highest LSE-C of 9.59 and the lowest LSE-D of 6.74.
This indicates that our unified audio-visual representation enables tight coupling between speech and facial motion.
In terms of video quality, JAVIS achieves comparable performance to the dedicated face generation models in the cascaded pipelines.
JAVIS also maintains natural speech quality, improving UTMOS and SpkSIM over the original FAC, which shows that the LoRA-adapted JRH preserves the pre-trained speech generation capability.

\noindent\textbf{Latency and generation speed.}
To evaluate the suitability for real-time interaction, we also measure the initial latency and real-time factor (RTF) of each system.
As shown in Table~\ref{tab:streaming}, JAVIS achieves both the lowest initial latency and the lowest RTF among all systems.
JAVIS starts responding within 1.27~s, 5.3$\times$ faster than FAC + FLOAT, and has 39\% lower initial latency than even FLOAT's streaming configuration.
The gap widens further against the other baselines: FAC + InfiniteTalk, FAC + Talker-T2AV, and Ex-Omni-2D require 252.0~s, 113.3~s, and 1210.7~s, respectively, before the response begins, and all run slower than real time (RTF~$>$~10).
In contrast, JAVIS generates faster than real time with an RTF of 0.541.
This efficiency stems from our end-to-end architecture, which generates motion chunk by chunk directly from the unified audio-visual representation without waiting for the full response speech, and operates on low-dimensional motion coefficients.
Notably, JAVIS matches the lip synchronization of FAC + InfiniteTalk while reducing initial latency by approximately 200$\times$.

\subsection{Subjective Evaluation}
We conduct a human evaluation to assess the perceptual quality of the generated responses, including whether the facial expressions are appropriate to the user's situation.
We sample 30 queries from VStyle with a balanced distribution across emotion categories.
A total of 26 raters listen to the user's speech and rate the response videos on a five-point scale in terms of audio-visual synchronization, motion naturalness, and facial expression appropriateness.

The mean scores are reported in Table~\ref{tab:mos} with 95\% confidence intervals.
As shown, JAVIS achieves the highest audio-visual synchronization score of 4.37, consistent with the objective evaluation results in Sec.~\ref{sec:objective}.
Its motion naturalness score of 3.92 is comparable to the best score of 4.02 obtained by FAC + InfiniteTalk, with overlapping confidence intervals.
Most importantly, JAVIS achieves the best facial expression appropriateness score of 3.76, clearly outperforming all baselines.
This suggests the effectiveness of our end-to-end architecture, in which the motion branch directly attends to the hidden states that encode the user's query context, whereas cascaded systems generate faces solely from the response speech or text.

\begin{table}[!t]
    \centering
    \caption{
    Subjective evaluation results. The results are presented with 95\% confidence intervals.
    }
    \label{tab:mos}
    \vspace{2mm}

    \small
    \setlength{\tabcolsep}{2pt}
    \renewcommand{\arraystretch}{1.12}

    \begin{tabular*}{\linewidth}{
        @{\extracolsep{\fill}}lccc@{}
    }
        \toprule
        \multirow{2}{*}{Method}
        & AV
        & Motion
        & Facial Expression \\
        & Sync $\uparrow$
        & Naturalness $\uparrow$
        & Appropriateness $\uparrow$ \\
        \midrule
        FAC + FLOAT          & 3.70{\scriptsize$\pm$0.15}             & 3.50{\scriptsize$\pm$0.15}             & 3.25{\scriptsize$\pm$0.15} \\
        FAC + InfiniteTalk   & \underline{4.25}{\scriptsize$\pm$0.11} & \textbf{4.02}{\scriptsize$\pm$0.11}    & \underline{3.44}{\scriptsize$\pm$0.14} \\
        FAC + Talker-T2AV    & 3.87{\scriptsize$\pm$0.13}             & 2.13{\scriptsize$\pm$0.12}             & 1.78{\scriptsize$\pm$0.11} \\
        Ex-Omni-2D           & 1.93{\scriptsize$\pm$0.13}             & 2.19{\scriptsize$\pm$0.13}             & 2.15{\scriptsize$\pm$0.14} \\
        \midrule
        \textbf{JAVIS} & \textbf{4.37}{\scriptsize$\pm$0.10}    & \underline{3.92}{\scriptsize$\pm$0.13} & \textbf{3.76}{\scriptsize$\pm$0.16} \\
        \bottomrule
    \end{tabular*}
\end{table}

\subsection{Ablation Study}
To validate the effect of each component, we report an ablation study in Table~\ref{tab:ablation}.
First, removing the lookahead representation $H_{t+1}$ degrades lip synchronization (LSE-C 9.59$\rightarrow$8.80, LSE-D 6.74$\rightarrow$7.55), indicating that access to upcoming speech context is important for generating accurate lip movements.
Removing JAVIS-Syn from dialogue adaptation also degrades lip synchronization (LSE-C 6.71, LSE-D 8.72) as well as speech quality (WER 1.21\%, UTMOS 4.03).
We attribute the lip-synchronization gains brought by JAVIS-Syn to its clean frontal faces and clearly visible lip movements, which provide more reliable supervision than existing dialogue videos that often contain low-quality frames with occlusion or blur.
Finally, the model trained only in the monologue stage shows degraded performance across all metrics, with particularly severe drops in speech intelligibility (WER 13.91\%) and lip synchronization (LSE-C 6.17, LSE-D 8.79), highlighting the importance of dialogue adaptation for audio-visual response generation.

\begin{table}[t]
    \centering
    \caption{
        Ablation study of JAVIS.
    }
    \label{tab:ablation}
    \vspace{2mm}

    \small
    \setlength{\tabcolsep}{3pt}
    \renewcommand{\arraystretch}{1.1}
    \resizebox{\columnwidth}{!}{%
        \begin{tabular}{lccccc}
            \toprule
            Model
            & IQ$\uparrow$
            & LSE-C$\uparrow$
            & LSE-D$\downarrow$
            & WER (\%)$\downarrow$
            & UTMOS$\uparrow$ \\
            \midrule
            \textbf{JAVIS}
            & \textbf{0.744}
            & \textbf{9.59}
            & \textbf{6.74}
            & \underline{0.44}
            & \underline{4.29} \\
            \quad w/o lookahead
            & \textbf{0.744} & \underline{8.80} & \underline{7.55} & \textbf{0.37} & \textbf{4.31} \\
            \quad w/o JAVIS-Syn
            & 0.737 & 6.71 & 8.72 & 1.21 & 4.03 \\
            \quad Stage 1 only
            & 0.734 & 6.17 & 8.79 & 13.91 & 3.74 \\
            \bottomrule
        \end{tabular}%
    }
\end{table}

\section{Conclusion}
We presented JAVIS, an end-to-end streaming framework for real-time audio-visual conversational agents.
Rather than animating a face after speech is synthesized, our framework generates speech and facial motion jointly from a single representation, allowing both modalities to be streamed together while sharing the dialogue context.
To address the scarcity of expressive audio-visual dialogue data, we further constructed JAVIS-Syn, a 334-hour synthetic corpus of emotionally rich multi-turn conversations and assistant-style interactions.
Extensive experiments demonstrate that JAVIS achieves strong audio-visual synchronization and facial expression appropriateness, while reducing initial latency by approximately 200$\times$ over a strong cascaded baseline and preserving the speech quality.
Future work includes extending JAVIS to full-duplex interaction with listening behavior.
We believe JAVIS offers a practical step toward conversational agents that engage with users through both voice and face in real time.

\vfill\pagebreak
\clearpage
\ninept
\bibliographystyle{IEEEbib}
\bibliography{refs}

\end{document}